\documentclass[default]{sn-jnl}

\usepackage{amsmath,amssymb,amsfonts} 
\usepackage{textcomp}
\usepackage{bm}
\usepackage{booktabs}
\usepackage{graphicx}
\usepackage{diagbox}
\usepackage{array}
\usepackage{subfigure}
\usepackage[ruled,linesnumbered]{algorithm2e}

\jyear{2022}%
\theoremstyle{thmstyleone}%
\theoremstyle{thmstyletwo}%

\theoremstyle{thmstylethree}%

\begin{document}

\title[ ]{Online Cross-Layer Knowledge Distillation on Graph Neural Networks with Deep Supervision}


\author[1]{\fnm{Jiongyu} \sur{Guo}}\email{jy.guo@zju.edu.cn}

\author[2]{\fnm{Defang} \sur{Chen}}\email{defchern@zju.edu.cn}

\author*[2]{\fnm{Can} \sur{Wang}}\email{wcan@zju.edu.cn}

\affil[1]{\orgdiv{College of Software}, \orgname{Zhejiang University}, \orgaddress{\city{Hangzhou} \country{China}}}

\affil[2]{\orgdiv{College of Computer Science and Technology}, \orgname{Zhejiang University}, \orgaddress{\city{Hangzhou} \country{China}}}


\abstract{Graph neural networks (GNNs) have become one of the most popular research topics in both academia and industry communities for their strong ability in handling irregular graph data. However, large-scale datasets are posing great challenges for deploying GNNs in edge devices with limited resources and model compression techniques have drawn considerable research attention. Existing model compression techniques such as knowledge distillation (KD) mainly focus on convolutional neural networks (CNNs). Only limited attempts have been made recently for distilling knowledge from GNNs in an offline manner. As the performance of the teacher model does not necessarily improve as the number of layers increases in GNNs, selecting an appropriate teacher model will require substantial efforts. To address these challenges, we propose a novel online knowledge distillation framework called \textit{Alignahead++} in this paper. Alignahead++ transfers structure and feature information in a student layer to the previous layer of another simultaneously trained student model in an alternating training procedure. Meanwhile, to avoid over-smoothing problem in GNNs, deep supervision is employed in Alignahead++ by adding an auxiliary classifier in each intermediate layer to prevent the collapse of the node feature embeddings. Experimental results on four datasets including PPI, Cora, PubMed and CiteSeer demonstrate that the student performance is consistently boosted in our collaborative training framework without the supervision of a pre-trained teacher model and its effectiveness can generally be improved by increasing the number of students. }




\maketitle

\section{Introduction} \label{int}
Recent years have witnessed the great success of graph neural networks (GNNs) in tackling irregular data and facilitating their downstream tasks such as node classification \cite{tomas2017semi, peter2018graph,william2017inductive}, link prediction \cite{kipf2016variational, peng2020graph}, and node clustering \cite{wang2019attributed}. However, the large-scale datasets and high-capacity architectures increase the computing and memory costs, 
making the deployment of large models difficult in platforms with limited resources. To address the above issues, mainstream model compression techniques such as knowledge distillation are explored and have achieved great success in compressing convolutional neural networks (CNNs) \cite{bucilua2006model,hinton2015distilling,romero2015fitnets,zagoruyko2017paying,chen2021cross,zhang2021confidence,chen2022simkd}. The most common form of knowledge distillation is offline KD where the compressed student model mimic the soft targets or hidden feature maps extracted from a well-trained teacher model, thus boosting the performance of student model and even outperforming the teacher model \cite{chen2021cross}. 
Comparing with CNNs, GNNs will encounter additional complexity in that besides soft targets or feature maps, it shall take into consideration graph topological structure in knowledge distillation.

\begin{figure*}[ht]
  \centering
  \includegraphics[width=0.9\linewidth]{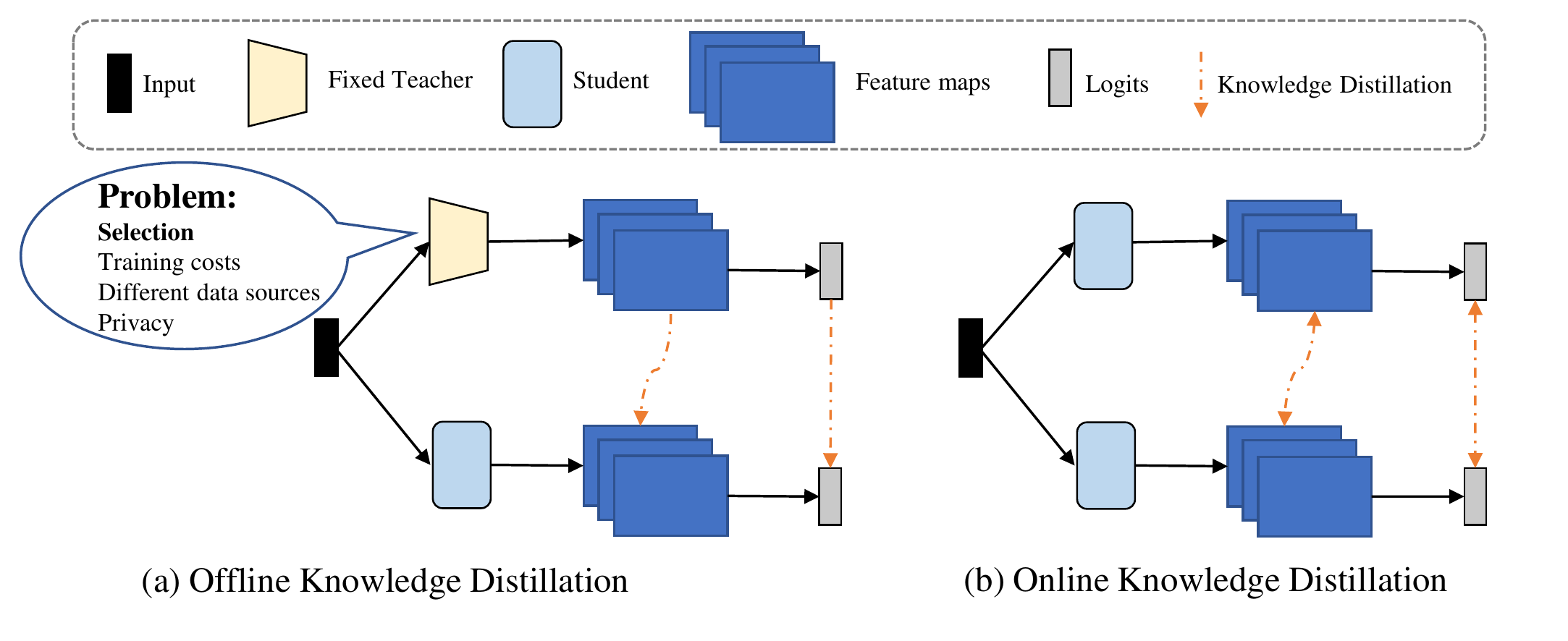}
  \caption{(a) Offline KD: the student model extracts knowledge from the pre-trained teacher model. (b) Online KD: two student models learn from each other in alternate training.}
  \label{fig:overview}
\end{figure*}

There exist only a few recent works in distilling knowledge from GNNs. Yang et al. \cite{yang2020distilling} modeled the local structure preserving (LSP) as the pairwise similarity of connected node features transferred from a pre-trained teacher model to a student model, which was the first attempt of knowledge distillation from GNNs. The subsequent researches mainly focus on adding auxiliary modules \cite{he2022compressing} to assist training, or introducing more effective knowledge \cite{joshi2021representation,yang2021extract} to capture comprehensive graph information, or graph-free settings \cite{deng2021graphfree}. Existing KD methods for GNNs are mostly offline (see Fig.\ref{fig:overview}(a)), i.e., the common teacher-student manner. An important part in offline knowledge distillation is to select a qualified teacher model. In CNNs, the teacher models are large models with more learning capacity and parameters. However, this does not apply to GNNs due to the \textit{over-smoothing} problem \cite{li2018deeper,Liu2020TowardsDG,Chen2020MeasuringAR}, i.e., the performance of GNNs does not necessarily improve as the number of layers increases. Selecting a qualified teacher model for student model is thus a time-consuming and laborious task as it requires substantial efforts and experiments. Furthermore, training costs, privacy and heterogeneous data sources between teacher and student are all issues that need to be addressed in distilling from GNNs.

Inspired by recent works that the student performance can be boosted by learning from other peers \cite{zhang2018deep,anil2018large,chen2020online}, as shown in Fig.\ref{fig:overview}(b), we propose a strategy called \textit{Alignahead++}, where each student model layer learns the structure information and feature information from the \textit{next layer} of another student model in an alternating training procedure. Theoretical analysis proves that Alignahead++ strategy with alternate training can spread structure and feature information over all layers of the two student models after several epochs and each student layer can effectively capture the global structure information. 
Meanwhile, previous studies have shown that with the increasing number of layers in GNNs, the over-smoothing problem will deteriorate significantly, i.e. the connected node embeddings will converge to a fixed point and thus degrade the performance of the model on downstream tasks. A disastrous training error will be back-propagated from the last layer to earlier layers in deep GNNs if only the last layer is supervised, which will disrupt the shallow feature embeddings. Therefore, we borrow the idea from the deep-supervised learning \cite{Szegedy2015GoingDW,Lee2015DeeplySupervisedN}, adding an auxiliary classifier to each intermediate layer of GNNs and then apply supervision to the predictions of each auxiliary classifier. 

One important issue in \textit{Alignahead++} is selecting the auxiliary classifier. Different from conventional data such as image data,  graph data needs to consider structure information in its feature embeddings. As Multilayer Perceptron (MLP) (commonly used in CNNs) cannot express the structure of the underlying graph data, we introduce structure-aware projection heads in Alignahead++, which uses a single GNN layer of the corresponding student model to capture the the graph structure. Experiments show that the effect of a single GNN layer is better than that of a single MLP layer, and only one layer is sufficient. Although our framework will increase a small number of parameters and time consumption, this is within an acceptable range. To verify the effectiveness of Alignahead++, we conduct experiments on different public datasets and different GNN architectures. The main contributions of this paper are highlighted as follows:

\begin{itemize}
\item We propose an online knowledge distillation framework called Alignahead++ for GNNs which is combined with the deep-supervised learning to supervise each intermediate layer. 
\item Each student layer is aligned ahead with the layer in different depth of another student model, which helps to spread structure and feature information over all layers after a couple of training iterations.
\item The introduced deep supervision module posing label supervision on intermediate layers alleviates the over-smoothing problem in deep GNNs.
\item Experimental results on various datasets and model architectures demonstrate that Alignahead++ framework can effectively improve the performance of student models.
\end{itemize}

The earlier version of this paper \cite{Guo2022Alignahead} was accepted by 2022 International Joint Conference on Neural Network (IJCNN 2022). In this work, we combine the Alignahead strategy with deep-supervised learning, termed as Alignahead++, to further improve the performance of the previously proposed online knowledge distillation framework and alleviate the over-smoothing problem in deep GNNs. In addition, we conduct a series of more detailed experiments, such as (a) two student models with different structures; (b) different layer numbers and structures (GNNs or MLPs) of auxiliary classifier and (c) sensitivity analysis of new hyper-parameters.

The rest of this paper is organized as follows. We introduce the related work of deep-supervised learning and knowledge distillation in GNNs in section 2. In Section 3, we introduce our proposed Alignahead++ strategy. Section 4 shows the experimental results, such as the node classification results on four kinds of datasets, sensitivity analysis of hyper-parameters and the exploration of auxiliary classifiers. Finally, we conclude this paper in Section 5.

\section{Related Work} \label{work}
\subsection{Knowledge Distillation}
Knowledge distillation is one of the most popular techniques in model compression \cite{bucilua2006model}, which utilizes a powerful but cumbersome teacher model to boost a compressed student model.  Hinton et al. \cite{hinton2015distilling} first proposed the concept of knowledge distillation, trying to transfer the soft targets, i.e., the teacher model predictions to the student models. It's believed that the soft targets can capture the relationship among classes and apply regularization during the training. In addition to soft targets, feature maps in the intermediate layers are another form of knowledge \cite{romero2015fitnets,zagoruyko2017paying,chen2021cross,zhang2021confidence}. Romero et al. \cite{romero2015fitnets} first put forward FitNet to reduce the distance of feature embeddings between teacher models and student models. Agoruyko et al. \cite{zagoruyko2017paying} considered the attention mechanism and extracted attention maps instead of features. SemCKD \cite{chen2021cross} utilizes the attention mechanism to automatically assign adaptive intermediate features for student model. SimKD \cite{chen2022simkd} achieves incredible effect just by reusing the teacher's classifier.

However, training a large teacher model in advance is time-consuming and requires high computing resources. In a worse case, we can't even get a teacher model. The online knowledge distillation \cite{zhang2018deep,anil2018large,chen2020online} is proposed to address these issues, where a group of student models are trained simultaneously by aggregating and aligning their outputs. Compared with the offline knowledge distillation mentioned above, online knowledge distillation can still boost the performance of the student model when reducing the training costs. Deep mutual learning \cite{zhang2018deep} trained a group of student models collaboratively by learning the soft targets from each other. OKDDip \cite{chen2020online} introduced multiple auxiliary peers and one team leader, proposing a novel two-level online knowledge distillation framework.

\subsection{Graph Neural Networks}
The recently proposed graph neural networks \cite{tomas2017semi,peter2018graph,william2017inductive,johannes2019predict,felix2019simplifying,chen2020simple} have achieved great success in processing irregular data, such as 3D point clouds \cite{Shi2020PointGNNGN} and protein property \cite{Shen2021NPIGNNPN}. Kipf et al. \cite{tomas2017semi} proposed the Graph Convolution Network (GCN), which propagates node features by spectral graph convolutions. GraphSAGE \cite{william2017inductive} further improved the scalability by sampling neighbor nodes and aggregating their features. Graph Atttention Network (GAT) \cite{peter2018graph} performs attention-based message passing to upgrade graph representations. 

However, as the number of layers increases, GNNs face the challenge of over-smoothing, that is, deep node feature embeddings will converge to a stationary point and become indistinguishable, which will degrade the performance of the model on downstream tasks. Some researchers try to design GNNs that can memorize initial node features to alleviate this problem, such as DeeperGCN \cite{Li2020DeeperGCNAY} and GCNII \cite{Chen2020SimpleAD}. Our work is perpendicular to DeeperGCN and other work mentioned above, that is, Alignahead++ can be used in any form of GNN models.

\subsection{Deeply-supervised learning}
Deeply supervised networks (DSNs) \cite{Lee2015DeeplySupervisedN} which added extra supervision to the hidden layer of convolutional neural networks (CNNs) were proposed to speed up the convergence of CNNs and solve the vanishing gradient problem. This technique has shown good ability in various tasks, such as image classification \cite{Lee2015DeeplySupervisedN}, machine translation \cite{Huang2021NonAutoregressiveTW} and image retargeting\cite{Mei2021DeepSI}. 

Zhang et al. \cite{zhang2019byot} proposed the self-distillation with the idea of deep supervision, where a single student model uses its deep predictions, feature maps and label to supervise its shallow layers. In GNNs, there is little research on deep supervision applications. Elinas et al. \cite{Elinas2022AddressingOI} made the first attempt to combine deep supervision with GNNs.
\subsection{Knowledge Distillation on Graph Neural Networks}
Knowledge distillation has achieved great success in computer vision and natural language processing tasks \cite{chen2022simkd,Jiao2020TinyBERTDB}. However, due to the particularity of graph data, that is, both feature embeddings and topological structures need to be considered, existing knowledge distillation methods cannot be directly applied to GNNs.

LSP \cite{yang2020distilling} was the first attempt of knowledge distillation on GNNs. It aligned the local structure of the student nodes with the corresponding teacher nodes based on the kernel function. Yang et al. \cite{yang2021extract} proposed a well-designed student model as a combination of label propagation and feature transformation. Joshi et al. \cite{joshi2021representation} extented the LSP module to global structure preservation (GSP) module and combined it with contrastive learning. GraphAKD \cite{he2022compressing} adaptively reduces the discrepancy between student and teacher models by training an additional generator and discriminator.

The work in this paper, Alignahead++, combines deep supervision with Alignahead, which alleviates the over-smoothing problem of the deep GNNs and can further improve the performance of shallow GNNs.

\section{Method} \label{work}
This section is organized into three subsections. Section 3.1 gives a brief introduction of local structure preserving. In section 3.2, we will detail our proposed \textit{Alignahead++}. In the last section, we will explain why the structure and feature information will flow from layer to layer.
\subsection{Local Structure Preserving (LSP)}
The core idea of LSP \cite{yang2020distilling} is to force the nodes of the student model to imitate the local structure of the teacher nodes. The similarity $L_{i,j}$ between node $i$ and its neighbor nodes $j\left(i,j\right)\in \varepsilon$ in the feature space is calculated with one of the following kernel functions and then normalized with a softmax function 

\begin{equation}
D\left(z_{i},z_{j}\right) = 
\begin{cases}
\left\|z_{i}-z_{j} \right\|_{2}^{2} & Euclidean \\ 
z_{i}\cdot z_{j} & Linear \\
\left(z_{i}\cdot z_{j}+c \right)^{d} & Poly\\
e^{-\frac{1}{2\sigma }\left\|z_{i}-z_{j} \right\|^{2}} & RBF
\end{cases} 
\end{equation}

\begin{equation}
L_{i,j} = \frac{e^{D\left(z_{i},z_{j}\right)}}{\sum_{\left(i,j\right)\in \varepsilon\left(e^{D\left(z_{i},z_{j}\right)}\right)}}
\end{equation}

where $z_{i}$ and $z_{j}$ are the node features. The local structure $L_{i}$ of node $i$ is quantified as the probability distribution of similarity between node $i$ and its neighbor nodes. 

The student model is trained with KL-divergence to mimic the local structure of the certain teacher layer in the feature space:
\begin{equation}
\mathcal{L}_{lsp} = \frac{1}{N}\sum_{i=1}^{N}D_{KL}\left(L_{i}^{T}||L_{i}^{S}\right).
\label{eq:lsp}
\end{equation}

\begin{figure*}[ht]
\centering
\includegraphics[width=0.95\linewidth]{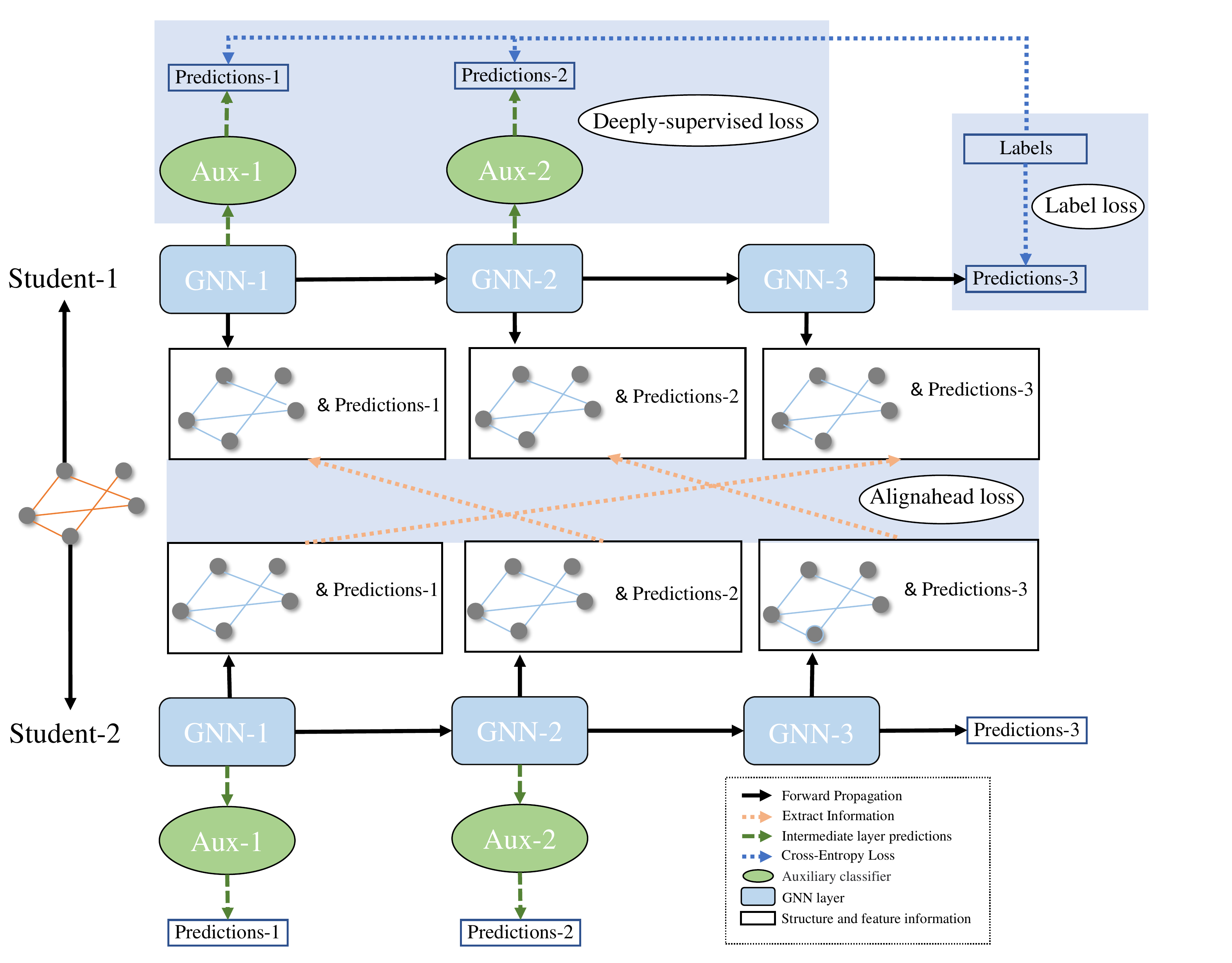}
\caption{Illustration of two alternately trained student models. The Alignahead++ strategy combines Alignahead with the deep supervision technique. It retains Alignahead's ability to transfer information between all layers and further imposes supervision in the intermediate layer to improve the model performance and alleviate the over-smoothing of deep GNNs. The loss function consists of three parts: the Label loss between model predictions and labels,the Alignahead loss, and the deeply-supervised loss.}
\label{fig:architecture}
\end{figure*}

\subsection{The proposed Alignahead++}
This section introduces the proposed Alignahead++ detailedly. We will start by explaining the motivation behind the previous Alignahead strategy. Although the method based on offline knowledge distillation has achieved some success in GNNs, the pre-training process of the teacher model faces various challenges. Online knowledge distillation eliminates the tedious pre-training process of teacher model and improves performance by simultaneously training a group of student models. Besides, LSP aligns only one layer between the student model and the teacher model but wastes information on the other layers. Therefore, we take advantage of the alternating training of online knowledge distillation and propose Alignahead strategy, where one student layer is aligned ahead with the layer in different depth of the another student model. In each round of alternate training, the structure information of each student layer is transferred to the previous layer of another student model, and finally the structure information will spread over all layers.

However, Alignahead strategy only considers the structure information of graph data while ignoring the rich semantic information contained in node feature embeddings. The intermediate semantic information of teacher model is always transferred to student model as knowledge in CNNs, and has achieved great success \cite{romero2015fitnets,zagoruyko2017paying,ahn2019variational,Yue2020MatchingGD, chen2021cross}. In Alignahead, the similarity between node feature embeddings calculated by kernel function only contains the relationship strength of two nodes, but the semantic information of each node itself is lost.
Therefore, we add an additional auxiliary classifier in each intermediate layer to make full use of the semantic information containing in nodes by transferring classification results in different semantic spaces. For the auxiliary classifier, we use a single GNN layer just the same as the backbone network instead of MLP commonly used in CNNs, because GNN layer has structure-aware function compared with MLP. Besides, student models trained by Alignahead still suffers from over-smoothing problem, so we introduce deep supervision technique, where we add the supervision of label to the each intermediate layer predictions to prevent node feature embeddings in the deep layer from becoming indistinguishable. Our collaborative training framework with two student models are illustrated in Fig. \ref{fig:architecture} and contains three parts of loss: Label loss, Deeply-supervised loss and Alignahead loss. Next, we'll talk about these three types of losses.

\subsubsection{Label loss}
This part of loss is conventional, that is, label based supervision is carried out on the backbone network predictions. Suppose we have two student models $S_{1}$ and $S_{2}$ with the same architecture, the intermediate layer number is $H$. Then the ground-truth loss of $S_{1}$ and $S_{2}$ is
\begin{equation}
\begin{split}
\mathcal{L}_{ce}^{S_{1}} = CrossEntropy({p}_{H}^{S_{1}},y)\\
\mathcal{L}_{ce}^{S_{2}} = CrossEntropy({p}_{H}^{S_{2}},y)
\end{split}
\label{eq:gdloss}
\end{equation}
where $\mathcal{L}_{ce}^{S_{1}/S_{2}}$ is a cross entropy loss function calculated with the student predictions $p_{H}^{S_{1}/S_{2}}$ and the label $y$.

\subsubsection{Deeply-supervised loss}
Deep supervision introduces the label information from the dataset to each layer of the backbone network directly by adding an additional auxiliary classifier in all intermediate layers. The auxiliary classifier is chosen a single GNN layer with the same structure as the backbone network. 

Let $f_{i,j}^{S_{1}}$ and $f_{i,j}^{S_{2}}$ represent the feature embeddings of node $j$  in the layer $i$. Since the last layer of the student model itself is a classifier, there is no need to add an auxiliary classifier. Thus, the feature embeddings of the last layer are the final predictions of the model, that is
\begin{equation}
    f_{H}^{S_{1}/S{2}} = p_{H}^{S_{1}/S_{2}} 
\end{equation}

\begin{algorithm}[H]
\caption{Alignahead++ for collaborative distillation}
\KwIn{The parameters $\theta _{1}$ and $\theta _{2}$ of the student $S_{1}$ and $S_{2}$ with auxiliary classifiers; The calculated local structure and intermediate layer predictions;hyper-parameter $\alpha,\beta,\lambda$ and label $y$.}

\KwOut{$S_{1}$ and $S_{2}$ with excellent performance.}

Initialization parameters $\theta _{1}$ and $\theta _{2}$.

\While{$epochs \le max\_epoch$}
{

    // Training loss of $S_{1}$ 
    
    Obtain the structure loss $\mathcal{L}_{str}^{S_{1}}$ with Equ. (\ref{eq:strloss}) and Equ. (\ref{eq:strloss1}).
    
    Obtain the feature loss $\mathcal{L}_{fea}^{S_{1}}$ with Equ. (\ref{eq:fealoss}) and Equ. (\ref{eq:fealoss1}).
    
    Obtain the deeply-supervised loss $\mathcal{L}_{ds}^{S_{1}}$ with Equ. (\ref{eq:dsloss}).
    
    Obtain the label loss $\mathcal{L}_{ce}^{S_{1}}$ with Equ. (\ref{eq:gdloss}).
         
    Construct the total loss for $S_{1}$ as Equ. (\ref{eq:ds-fea1}) and Equ. (\ref{eq:finalloss1}).
    
    \textbf{Update} the parameter $\theta _{1}$ while keeping $\theta _{2}$ fixed.
    
    // Training loss of $S_{2}$ 
    
    Obtain the structure loss $\mathcal{L}_{str}^{S_{2}}$ with Equ. (\ref{eq:strloss}) and Equ. (\ref{eq:strloss2}).
    
    Obtain the feature loss $\mathcal{L}_{fea}^{S_{2}}$ with Equ. (\ref{eq:fealoss}) and Equ. (\ref{eq:fealoss2}).
    
    Obtain the deeply-supervised loss $\mathcal{L}_{ds}^{S_{2}}$ with Equ. (\ref{eq:dsloss}).
        
    Obtain the label loss $\mathcal{L}_{ce}^{S_{2}}$ with Equ. (\ref{eq:gdloss}).
         
    Construct the total loss for $S_{1}$ as Equ. (\ref{eq:ds-fea2}) and Equ. (\ref{eq:finalloss2}).

    \textbf{Update} the parameter $\theta _{2}$ while keeping $\theta _{1}$ fixed.
        
}
\end{algorithm}

For the other layers, feature embeddings need to be mapped into the class semantic space through their respective auxiliary classifiers. We have tried two types of auxiliary classifier, i.e., MLP and one single GNN layer. During the training process, the auxiliary classifiers and the backbone network are updated simultaneously. Compared with MLP, GNN layer can capture the topology information (including the neighbor structure of the nodes), that is, 
\begin{equation}
    p_{i,j}^{S_{1}/S_{2}} = \left\{
\begin{array}{ll}
GNN\left(f_{i,j}^{S_{1}/S_{2}},f_{i,k}^{S_{1}/S_{2}} \right), \left(j,k\right)\in \varepsilon \\
MLP\left(f_{i,j}^{S_{1}/S_{2}}\right)
\end{array}
\right.
\end{equation}
where $p_{i,j}^{S_{1}/S_{2}}$ are the predictions of node $j$ in layer $i$ of two student models. In addition, it is natural and reasonable to use classifiers consistent with the backbone network structure. For the above reasons, GNN layer is more suitable as the auxiliary classifier. But anyway, we explore the performance of these two classifiers and the effect of the layer numbers in the experiment section.

Similar with the label loss, the target of deeply-supervised loss is the intermediate layer predictions. In order to simplify the form, the last-layer predictions are still added in the calculation of deeply-supervised loss.

\begin{equation}
\begin{split}
\mathcal{L}_{ds}^{S_{1}} =  \sum_{i=1}^{H}CrossEntropy({p}_{i}^{S_{1}},y)\\
\mathcal{L}_{ds}^{S_{2}} = 
\sum_{i=1}^{H}CrossEntropy({p}_{i}^{S_{2}},y)
\end{split}
\label{eq:dsloss}
\end{equation}

\subsubsection{Alignahead loss}
Alignahead loss is divided into two parts, namely, structure loss and feature loss. Structure loss is used to capture the relationship between nodes while feature loss is used to supervise the semantic information of nodes themselves. We will start by introducing the structure loss.

For two student models $S_{1}$ and $S_{2}$ with $H$ layers, we use $l_{i,j}^{S_{1}}$ and $l_{i,j}^{S_{2}}$ to represent the local structure of node $j$ in the $i$-th layer. As mentioned above, the local structure $l_{i,j}^{S_{1}/S_{2}}$ is quantified as the probability distribution of similarity between node $i$ and its neighbor nodes in layer $j$. For $S_{1}$, its local structure of $i$-th layer is required to match the $\left(i+1\right)$-th layer of $S_{2}$ and the final layer $H$ is required to match the first layer. The structure loss is calculated as the sum of KL divergence loss over all layers
\begin{equation}
\mathcal{L}_{str}^{S_{1}} = \sum_{i=1}^{H}L^{S_{1}}_{{l}_{i}},\quad
\mathcal{L}_{str}^{S_{2}} = \sum_{i=1}^{H}L^{S_{2}}_{{l}_{i}},
\label{eq:strloss}
\end{equation}
where 
\begin{equation}
{L}^{S_{1}}_{{l}_{i}} = \frac{1}{N}\sum_{j=1}^{N}D_{KL}\left(l_{i+1,j}^{S_{2}}||l_{i,j}^{S_{1}}\right),\label{eq:strloss1}
\end{equation}
\begin{equation}
{L}^{S_{2}}_{{l}_{i}} = \frac{1}{N}\sum_{j=1}^{N}D_{KL}\left(l_{i+1,j}^{S_{1}}||l_{i,j}^{S_{2}}\right).\label{eq:strloss2}
\end{equation}

Next it comes to feature loss. Similar to the structure loss, feature loss also adopts Alignahead strategy. We align the predictions $p_{i}^{S_{1}}$ of $S_{1}$ in layer $i$ with $p_{i+1}^{S_{2}}$ of $S_{2}$ in layer $i+1$ and take the sum of KL divergence of all layers as the feature loss
\begin{equation}
\mathcal{L}_{fea}^{S_{1}} =  \sum_{i=1}^{H}L^{S_{1}}_{{p}_{i}},\quad
\mathcal{L}_{fea}^{S_{2}} =  \sum_{i=1}^{H}L^{S_{2}}_{{p}_{i}},
\label{eq:fealoss}
\end{equation}
where 
\begin{equation}
{L}^{S_{1}}_{{p}_{i}} = \frac{1}{N}\sum_{j=1}^{N}D_{KL}\left(p_{i+1,j}^{S_{2}}||p_{i,j}^{S_{1}}\right),\label{eq:fealoss1}
\end{equation}
\begin{equation}
{L}^{S_{2}}_{{p}_{i}} = \frac{1}{N}\sum_{j=1}^{N}D_{KL}\left(p_{i+1,j}^{S_{1}}||p_{i,j}^{S_{2}}\right).\label{eq:fealoss2}
\end{equation}

\subsubsection{The combination of loss}
We will talk about how to get the final loss function in this section. For the intermediate layer predictions of the student model, they are constrained by both the labels and the previous layer of the peer model, which can correspond to labels and soft targets in the offline knowledge distillation. Therefore, we use the hyper-parameter $\lambda$ to balance the two parts of loss
\begin{equation}
\mathcal{L}_{ds-fea}^{S_{1}} = \lambda \mathcal{L}_{fea}^{S_{1}}+(1-\lambda )\mathcal{L}_{ds}^{S_{1}}
\label{eq:ds-fea1}
\end{equation}
\begin{equation}
\mathcal{L}_{ds-fea}^{S_{2}} = \lambda \mathcal{L}_{fea}^{S_{2}}+(1-\lambda )\mathcal{L}_{ds}^{S_{2}}
\label{eq:ds-fea2}
\end{equation}
Then combine it with label loss and structure loss
\begin{equation}
\mathcal{L}^{S_{1}} = (1-\beta )\mathcal{L}_{ce}^{S_{1}}+\frac{\beta}{H} \mathcal{L}_{ds-fea}^{S_{1}}+\alpha \mathcal{L}_{str}^{S_{1}}
\label{eq:finalloss1}
\end{equation}
\begin{equation}
\mathcal{L}^{S_{2}} = (1-\beta )\mathcal{L}_{ce}^{S_{2}}+\frac{\beta}{H} \mathcal{L}_{ds-fea}^{S_{2}}+\alpha \mathcal{L}_{str}^{S_{2}}
\label{eq:finalloss2}
\end{equation}
where $\lambda,\beta \in [0,1]$. The second loss term is divided by the number of layers $H$ to match the magnitude of label loss. We fix $\alpha=1$, since we have found in our previous work \cite{Guo2022Alignahead} that the experiment results are not sensitive to the value of this hyper-parameter. In order to show the role of the other two hyper-parameters more clearly, Table \ref{tab:extreme value} shows the meanings of the first two loss terms when they take extreme values. In the experiment section, we will also conduct sensitivity studies on these two hyper-parameters. Two student models $S_{1}$ and $S_{2}$ are trained alternately as shown in Algorithm 1.

\begin{table}[ht]
\centering
\renewcommand{\arraystretch}{1.0}
\caption{The meanings of the first two loss terms when $\lambda$ and $\beta$ take extreme values.}
\resizebox{0.7\textwidth}{!}{
\begin{tabular}{ccc}
\toprule
\multicolumn{1}{c}{\diagbox{$\beta$}{$\lambda$}} & 0 & 1 \\ 
\midrule
\multicolumn{1}{c|}{0}  & Label loss & Label loss  \\
\multicolumn{1}{c|}{1}  & Deeply-supervised loss & Feature loss \\ 
\bottomrule
\end{tabular}
}
\label{tab:extreme value}
\end{table}

\subsection{Why the structure and feature information spread over all layers?}

\begin{figure*}[ht]
  \centering
  \includegraphics[width=0.9\textwidth]{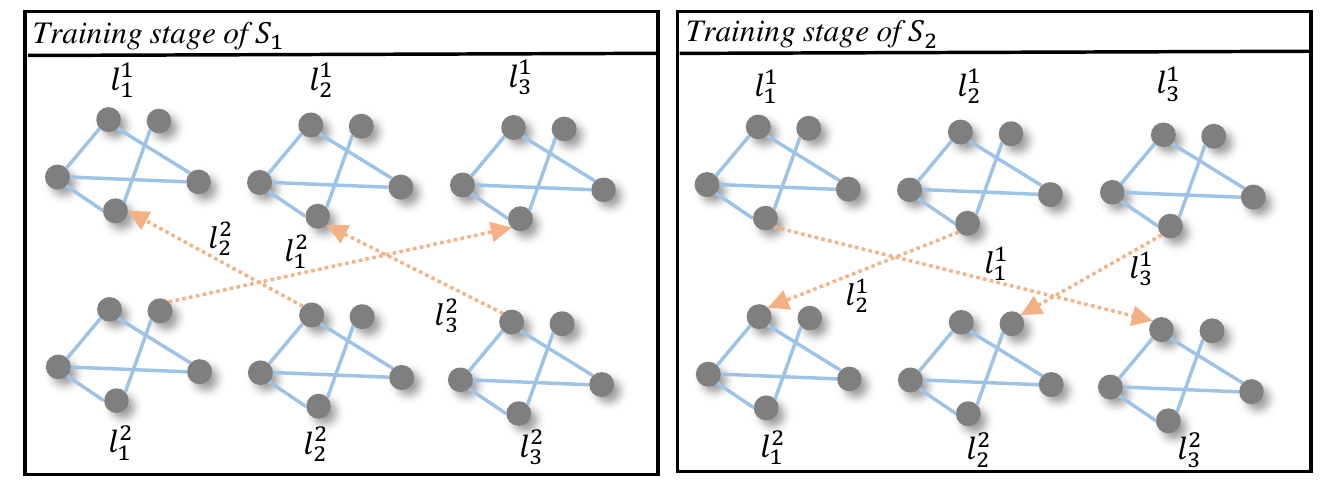}
  \caption{The arrows represent the information flow in alternating training of two student models.}
  \label{fig:flow}
\end{figure*}

\begin{table*}[ht]
\centering
\caption{After six iterations, each student layer captures the structure information from all the other layers. One iteration includes the alternating training of Student-1 and student-2}
\renewcommand{\arraystretch}{1}
\resizebox{0.9\textwidth}{!}{%
\begin{tabular}{lllllllllllllll}
\toprule
\multicolumn{2}{c}{Iteration number} & ~~~0 & \multicolumn{2}{c}{1} & \multicolumn{2}{c}{2} & \multicolumn{2}{c}{3} & \multicolumn{2}{c}{4} & \multicolumn{2}{c}{5} & \multicolumn{2}{c}{6} \\ 
\midrule
\multicolumn{2}{c}{Training stage} & Initial & $S_{1}$ & \multicolumn{1}{c}{$S_{2}$} & $S_{1}$ & \multicolumn{1}{c}{$S_{2}$} & $S_{1}$ & \multicolumn{1}{c}{$S_{2}$} & $S_{1}$ & \multicolumn{1}{c}{$S_{2}$} & $S_{1}$ & \multicolumn{1}{c}{$S_{2}$} & $S_{1}$ & \multicolumn{1}{c}{$S_{2}$} \\
\midrule
Student-1 & Layer-1 & \multicolumn{1}{l|}{\bm{~~$l_{1}^{1}$}}& \bm{$l_{2}^{2}$} & \multicolumn{1}{l|}{$l_{2}^{2}$} &  \bm{$l_{3}^{1}$}& \multicolumn{1}{l|}{$l_{3}^{1}$} &  \bm{$l_{1}^{2}$}& \multicolumn{1}{l|}{$l_{1}^{2}$} &  \bm{$l_{2}^{1}$}& \multicolumn{1}{l|}{$l_{2}^{1}$} &  \bm{$l_{3}^{2}$}& \multicolumn{1}{l|}{$l_{3}^{2}$} &  \bm{$l_{1}^{1}$}& $l_{1}^{1}$ \\ \cmidrule(l){2-15}
& Layer-2 & \multicolumn{1}{l|}{~~$l_{2}^{1}$} & $l_{3}^{2}$ & \multicolumn{1}{l|}{$l_{3}^{2}$} & $l_{1}^{1}$ & \multicolumn{1}{l|}{$l_{1}^{1}$} &  $l_{2}^{2}$& \multicolumn{1}{l|}{$l_{2}^{2}$} &  $l_{3}^{1}$& \multicolumn{1}{l|}{$l_{3}^{1}$} &  $l_{1}^{2}$& \multicolumn{1}{l|}{$l_{1}^{2}$} &  $l_{2}^{1}$& $l_{2}^{1}$ \\ \cmidrule(l){2-15}
& Layer-3 & \multicolumn{1}{l|}{~~$l_{3}^{1}$} & $l_{1}^{2}$ & \multicolumn{1}{l|}{$l_{1}^{2}$} &  $l_{2}^{1}$& \multicolumn{1}{l|}{$l_{2}^{1}$} &  $l_{3}^{2}$& \multicolumn{1}{l|}{$l_{3}^{2}$} &  $l_{1}^{1}$& \multicolumn{1}{l|}{$l_{1}^{1}$} &  $l_{2}^{2}$& \multicolumn{1}{l|}{$l_{2}^{2}$} &  $l_{3}^{1}$& $l_{3}^{1}$ \\ \midrule
Student-2 & Layer-1 & \multicolumn{1}{l|}{~~$l_{1}^{2}$} & $l_{1}^{2}$ & \multicolumn{1}{l|}{$l_{2}^{1}$} &  $l_{2}^{1}$& \multicolumn{1}{l|}{$l_{3}^{2}$} &  $l_{3}^{2}$& \multicolumn{1}{l|}{$l_{1}^{1}$} &  $l_{1}^{1}$& \multicolumn{1}{l|}{$l_{2}^{2}$} &  $l_{2}^{2}$& \multicolumn{1}{l|}{$l_{3}^{1}$} & $l_{3}^{1}$ & $l_{1}^{2}$ \\ \cmidrule(l){2-15} 
 & Layer-2 & \multicolumn{1}{l|}{~~$l_{2}^{2}$} & $l_{2}^{2}$ & \multicolumn{1}{l|}{$l_{3}^{1}$} &  $l_{3}^{1}$& \multicolumn{1}{l|}{$l_{1}^{2}$} &  $l_{1}^{2}$& \multicolumn{1}{l|}{$l_{2}^{1}$} &  $l_{2}^{1}$& \multicolumn{1}{l|}{$l_{3}^{2}$} &  $l_{3}^{2}$& \multicolumn{1}{l|}{$l_{1}^{1}$} & $l_{1}^{1}$ & $l_{2}^{2}$ \\ \cmidrule(l){2-15} 
 & Layer-3 &  \multicolumn{1}{l|}{~~$l_{3}^{2}$} & $l_{3}^{2}$ & \multicolumn{1}{l|}{$l_{1}^{1}$} &  $l_{1}^{1}$& \multicolumn{1}{l|}{$l_{2}^{2}$}&  $l_{2}^{2}$&  \multicolumn{1}{l|}{$l_{3}^{1}$}& $l_{3}^{1}$&  \multicolumn{1}{l|}{$l_{1}^{2}$}&  $l_{1}^{2}$&  \multicolumn{1}{l|}{$l_{2}^{1}$}& $l_{2}^{1}$ & $l_{3}^{2}$ \\ 
\bottomrule
\end{tabular}%
}
\label{tab: alignahead}
\end{table*}
This section will explain why the structure and feature information spreads over all layers. For convenience, we will only focus on structure information, and it's the same with feature information. Fig. \ref{fig:flow} reveals the flow of structure information during an alternating training procedure. 
It can be seen that each student layer is aligned ahead with the next layer of another student model and the last layer is to match the first layer specially. 
In this way, the two student models exchange structure information with each other and propagate it in different layer depth. 

Table \ref{tab: alignahead} shows the distribution of structure information of two student models with three hidden layers in six iterations. 
Theoretically, the structure information of each layer is circulated once inside the two student models. 
Take the layer-1 of the Student-1 model as an example, it gathers the structure information in $l_{2}^{2}$, $l_{3}^{1}$, $l_{1}^{2}$, $l_{2}^{1}$, $l_{3}^{2}$ and $l_{1}^{1}$ one-by-one, and returns to the initial state in the sixth iteration. That is to say, the layer-1 of the Student-1 model indeed collects the structure information from all layers of two student models within the first six iterations. Similar phenomenon also happens in other layers, making the structure information spread over all layers.

\section{Experiments}
We first introduce four benchmark datasets, the experimental settings, three popular graph neural network architectures and compared methods adopted in our experiments in section \ref{subsec:datasets-models}. Section \ref{subsec:ppi} and \ref{subsec:cite} detail the experimental results with different methods in four datasets and demonstrate that Alignahead++ can alleviate the over-smoothing problem in deep GNNs. Next, We explore the performance of the student models with different hyper-parameters choice and two different architectures in Sections \ref{subsec:sensitivity} and \ref{subsec:difstu}, respectively. Finally, we explore the effect of auxiliary classifiers and student numbers.

\subsection{Datasets, model architectures and compared methods}
\label{subsec:datasets-models}
Four benchmark datasets are used in our experiments, and their summary are shown in Table \ref{tab:dataset} and listed as follows:
\begin{itemize}
\item PPI \cite{marinka2017predicting} consists of 24 graphs corresponding to different human tissues, where 20 graphs are used for training, two graphs are used for validation, and another 2 graphs are used for testing. The average node numbers for each graph is 2372. Each node has 50 features, consisting of positional gene sets, motif gene sets, and immunological signatures. The number of labels is 121 and a node can have multiple labels at the same time.
\item Cora \cite{bojchevski2017deep,McCallum2004AutomatingTC} and CiteSeer \cite{Sen2008CollectiveCI} consist of citation networks on computer science and both two are widely used in GNNs. Each node represents a paper, and the edge directly connected between nodes represents the citation relationship. The feature is a binary vector indicating the presence or absence of a word in the paper, and the label represents the topic.
\item PubMed \cite{Namata2012QuerydrivenAS} is the citation network on diabetes in PubMed database. The meanings of nodes and edges are the same as Cora. Node features are TF/IDF-weighted word frequencies and the label represents the type of diabetes.
\end{itemize}

\begin{table*}[t]
\centering
\renewcommand{\arraystretch}{1.1}
\caption{Summary of the datasets.}
\resizebox{1\linewidth}{!}{
\begin{tabular}{cccccccccc}
\toprule
\textbf{Datasets} & \textbf{\#Graphs} & \textbf{\#Nodes} & \textbf{\#Edges} & \textbf{\#Features} & \textbf{\#Classes} & \textbf{\#Training} &\textbf{\#Validation} &\textbf{\#Test} &\textbf{\#Metric} \\ 
\midrule
PPI             & 24       & 56944   & 818716  & 50         & 121 (multilabel) & 20 (Graphs) & 2 (Graphs) & 2 (Graphs)  &F1 score \\
Cora     & 1        & 2708   & 5429   & 1433       & 7             &
140 (Nodes) & 500 (Nodes) & 1000 (Nodes)  & Accuracy \\
CiteSeer    & 1        & 3327   & 4732  & 3703       & 6       &
120 (Nodes) & 500 (Nodes) & 1000 (Nodes)        & Accuracy \\
PubMed & 1        & 19717   & 44338  & 500        & 3          &
60 (Nodes) & 500 (Nodes) & 1000 (Nodes)     & Accuracy \\
\bottomrule
\end{tabular}
}
\label{tab:dataset}
\end{table*}

The adopted models are briefly listed as follows:
\begin{itemize}
\item GCN \cite{tomas2017semi} propagates node features by defining spectral graph convolutions on graph data. It is the most widely used baseline model in GNNs.
\item SAGE \cite{william2017inductive} updates node features by sampling neighbor nodes and aggregating their information. SAGE-mean and SAGE-pool are used as student models.
\item GAT \cite{peter2018graph} introduces the attention mechanism to automatically assign weights to nodes' neighbors. In the PPI dataset, we adopt GAT with different layers and feature dimensions.
\end{itemize}

The results of three methods are presented in comparison:
\begin{itemize}
\item \textit{Baseline} means the student model is trained with the labels alone.
\item \textit{Alignahead} is our previous work proposed specifically for GNNs \cite{Guo2022Alignahead}.
Each student model layer learns the structure information from the next layer of another student model in an alternating training procedure.
\item \textit{OC} is the ablation model of \textit{Alignahead}. It keeps the setting of online knowledge distillation but changes cross-layer matching to one-to-one correspondence, which means each layer of the student model is matched with the corresponding layer of the peer model.
\item \textit{Alignahead++} means the student models are trained with our proposed online cross-layer distillation framework combined with deep supervision, as shown in Algorithm 1.
\end{itemize}

In the case of the PPI dataset, we train all the three student models on the visible graph to make predictions on completely invisible graphs. We refer to the experiment settings of \cite{yang2020distilling}, where the learning rate, weight decay, optimizer and epochs are 0.005, 0, Adam and 300, respectively. We adopt the RBF kernel function for distance calculation in LSP module, and $\sigma$ is 100. 
While on other four datasets, we apply GCN \cite{tomas2017semi} and two variants of SAGE \cite{william2017inductive} as student models to predict the unknown nodes on the visible graph. 
We adopt one set of experiment settings from \cite{william2017inductive}, where the learning rate, weight decay, optimize are 0.001, 0.0001, Adam, respectively, we also set up 300 epochs and the RBF kernel function is adopted for all models. For the hyper-parameters, we take $\alpha=1$, $\beta=0.4$ and $\lambda=0.2$. For the three methods of OC, Alignahead and Alignahead++, we only report the higher metric in the two student models. All our experiments are performed on an NVIDIA 2080Ti GPU. 

\subsection{Node classification on PPI}
\label{subsec:ppi}

On the PPI dataset, we use three GNN architectures, namely GAT,GCN and SAGE-pool. For each GNN, we take different layers and feature dimensions for experiments. The experimental results are shown in Table \ref{tab:ppi}.

\begin{table*}[ht]
\centering
\renewcommand{\arraystretch}{1.1}
\caption{The results of GAT, GCN and SAGE-pool student models on PPI. The layers of GAT represent the attention heads in each layer.}
\resizebox{1\linewidth}{!}{
\begin{tabular}{@{}ccc|c|c|ccc|ccc@{}}
\toprule
\multicolumn{3}{c|}{}                        & Baseline     & OC       & \multicolumn{3}{c|}{Alignahead} & \multicolumn{3}{c}{Alignahead++}  \\ \midrule
Model     & Layers            & Feature dimensions & F1-score & F1-score & Capacity   & Time   & F1-score  & Capacity & Time  & F1-score        \\ \midrule
GCN       & 2                 & 256          & 0.6153   & 0.6176   & 0.11M      & 0.68s  & 0.6183    & 0.17M    & 0.73s & \textbf{0.6245} \\
GCN       & 2                 & 512          & 0.6397   & 0.6599   & 0.37M      & 0.72s  & 0.6606    & 0.79M    & 0.76s & \textbf{0.6743} \\
GCN       & 3                 & 512          & 0.6163   & 0.6512   & 0.89M      & 0.85s  & 0.6715    & 1.45M    & 0.90s & \textbf{0.7050} \\
GCN       & 5                 & 512          & 0.6130   & 0.6865   & 0.53M      & 1.15s  & 0.7018    & 0.91M    & 1.25s & \textbf{0.7160} \\
GCN       & 10                & 512          & 0.4085   & 0.5280   & 2.38M      & 1.87s  & 0.5173    & 3.37M    & 2.06s & \textbf{0.6620} \\ \midrule
SAGE-pool & 3                 & 128          & 0.9018   & 0.9292   & 0.16M      & 0.67s  & 0.9316    & 0.26M    & 0.75s & \textbf{0.9393} \\
SAGE-pool & 3                 & 256          & 0.9582   & 0.9810   & 0.55M      & 0.71s  & 0.9818    & 0.81M    & 0.84s & \textbf{0.9851} \\
SAGE-pool & 4                 & 64           & 0.7798   & 0.8050   & 0.07M      & 0.67s  & 0.8056    & 0.13M    & 0.74s & \textbf{0.8284} \\
SAGE-pool & 4                 & 128          & 0.8848   & 0.9300   & 0.21M      & 0.69s  & 0.9338    & 0.36M    & 0.78s & \textbf{0.9495} \\
SAGE-pool & 4                 & 256          & 0.9540   & 0.9801   & 0.75M      & 0.82s  & 0.9815    & 1.13M    & 0.92s & \textbf{0.9863} \\ \midrule
GAT       & {[}3,3,3,3,3{]}   & 32           & 0.9090   & 0.9085   & 0.13M      & 0.83s  & 0.9105    & 0.34M    & 1.03s & \textbf{0.9140} \\
GAT       & {[}3,3,3,3,3{]}   & 64           & 0.9736   & 0.9757   & 0.37M      & 0.93s  & 0.9770    & 0.79M    & 1.18s & \textbf{0.9791} \\
GAT       & {[}3,3,3,3{]}     & 128          & 0.9820   & 0.9832   & 0.89M      & 1.02s  & 0.9836    & 1.45M    & 1.22s & \textbf{0.9873} \\
GAT       & {[}2,2,2,2,2{]}   & 128          & 0.9780   & 0.9796   & 0.53M      & 0.97s  & 0.9800    & 0.91M    & 1.15s & \textbf{0.9846} \\
GAT       & {[}2,2,2,2,2,2{]} & 256          & 0.9882   & 0.9891   & 2.38M      & 1.62s  & 0.9897    & 3.37M    & 1.93s & \textbf{0.9908} \\ \bottomrule
\end{tabular}
}
\label{tab:ppi}
\end{table*}

It can be seen that student models trained with our new proposed Alignahead++ consistently achieve better performance compared to those trained with labels or Alignahead strategy. The comparison between the results of one-to-one correspondence (OC) and Alignahead demonstrates Alignahead can indeed help the student layer learn global semantics by flowing structure information. By introducing feature information and combining with deep supervision, Alignahead++ can deliver more comprehensive and accurate information, further improving the performance of student models. Due to the addition of auxiliary classifiers, the model capacity and training time are increased. Nevertheless, knowledge distillation is generally aimed at small capacity models, which makes the increased costs affordable. 

In addition, we find that when the layer number of GCN reaches 10, Alignahead++ has the biggest improvement compared with Alignahead. It is probably due to the reason that GNNs suffer from over-smoothing problem, that is, the feature embeddings of nodes in deep layer become indistinguishable as the layer number increases. We believe that our Alignahead++ can alleviate the over-smoothing \cite{li2018deeper,Liu2020TowardsDG,Chen2020MeasuringAR} problem to some extent and delve more deeply in the next section. 

\begin{figure*}[ht]
\centering
\subfigure[Layers-3, heads-3, feature dimensions-128]{
\begin{minipage}[ht]{0.5\linewidth}
\centering
\includegraphics[width=0.9\linewidth]{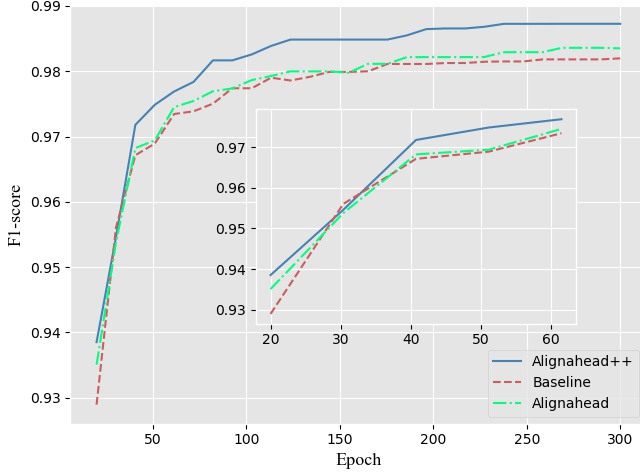}
\end{minipage}%
}%
\subfigure[Layers-4, heads-2, feature dimensions-128]{
\begin{minipage}[ht]{0.5\linewidth}
\centering
\includegraphics[width=0.9\linewidth]{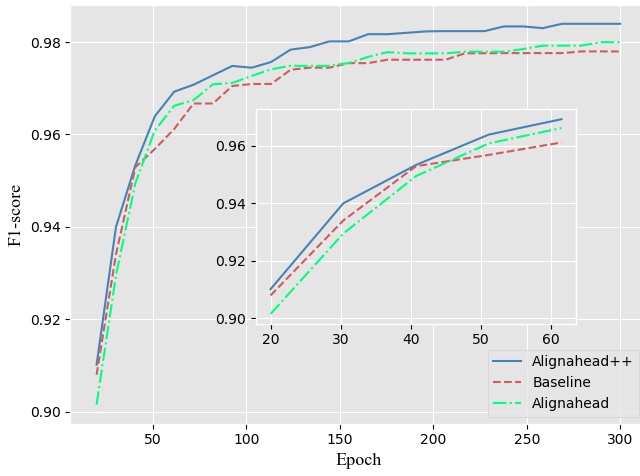}
\end{minipage}%
}%

\subfigure[Layers-4, heads-3, feature dimensions-64]{
\begin{minipage}[ht]{0.5\linewidth}
\centering
\includegraphics[width=0.9\linewidth]{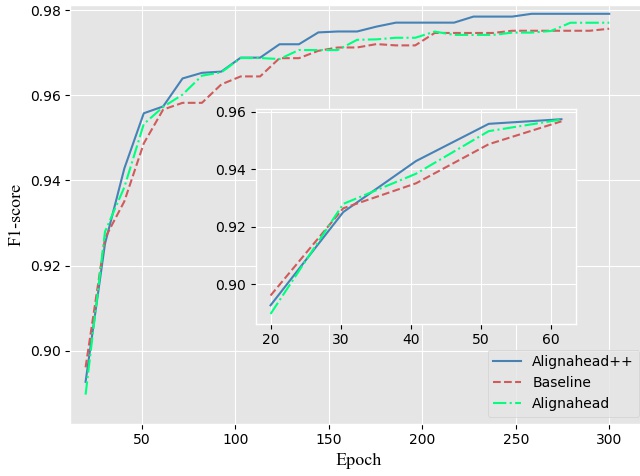}
\end{minipage}
}
\subfigure[Layers-5, heads-2, feature dimensions-256]{
\begin{minipage}[ht]{0.5\linewidth}
\centering
\includegraphics[width=0.9\linewidth]{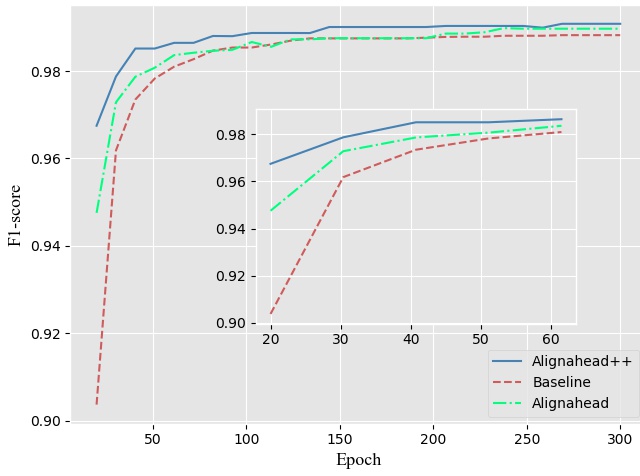}
\end{minipage}
}%
\centering
\caption{F1-score curves of different GAT models during training.The small images magnify the results of the first 50 epochs.}
\label{fig:curve}
\end{figure*}

Then, we explore the F1-score in GAT models with different structures under different methods during training, and the results are shown in Fig. \ref{fig:curve}. We start drawing from the F1-score of Alignahead++ greater than 0.9 without OC method to prevent curves from tangling together. It can be seen that our Alignahead++ always outperforms Alignahead and Baseline, demonstrating the superiority of Alignahead++. 

\begin{figure*}[ht]
\centering
\subfigure[GCN on Cora]{
\begin{minipage}[ht]{0.45\linewidth}
\includegraphics[width=1\linewidth]{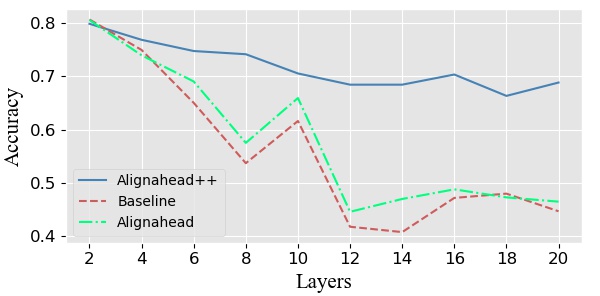}
\end{minipage}%
}%
\subfigure[SAGE-mean on Cora]{
\begin{minipage}[ht]{0.45\linewidth}
\includegraphics[width=1\linewidth]{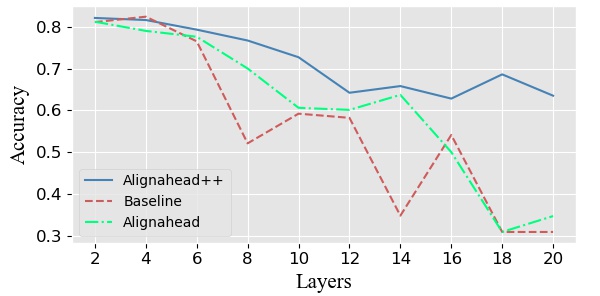}
\end{minipage}%
}%
\centering
\caption{The accuracy of GCN and SAGE-mean with different layers on Cora. The feature dimensions are 128.}
\label{fig:over-smoothing}
\end{figure*}

\subsection{Node classification on citation datasets}
\label{subsec:cite}

\begin{table*}[ht]
\centering
\renewcommand{\arraystretch}{1}
\caption{The results of different GNNs on citation datasets}
\resizebox{\linewidth}{!}{
\begin{tabular}{@{}lccc|c|c|clcclc@{}}
\toprule
\multicolumn{4}{c|}{}                              & Baseline          & OC             & \multicolumn{3}{c}{Alignahead}                          & \multicolumn{3}{c}{Alignahead++}  \\ \midrule
Dataset  & Model     & Layers & Feature dimensions & Acc            & Acc            & Capacity & Time   & \multicolumn{1}{c|}{Acc}            & Capacity & Time   & Acc            \\ \midrule
Cora     & GCN       & 3      & 128                & 0.772          & 0.770          & 0.22M    & 0.023s & \multicolumn{1}{c|}{0.776}          & 0.22M    & 0.031s & \textbf{0.785} \\
Cora     & GCN       & 10     & 128                & 0.616          & 0.614          & 0.33M    & 0.062s & \multicolumn{1}{c|}{0.659}          & 0.34M    & 0.075s & \textbf{0.705} \\
Cora     & SAGE-mean & 3      & 128                & \textbf{0.818} & \textbf{0.818} & 0.43M    & 0.026s & \multicolumn{1}{c|}{0.790}           & 0.44M    & 0.028s & 0.802          \\
Cora     & SAGE-mean & 10     & 128                & 0.592          & 0.593          & 0.67M    & 0.065s & \multicolumn{1}{c|}{0.606}          & 0.68M    & 0.088s & \textbf{0.727} \\
Cora     & SAGE-pool & 3      & 128                & 0.706          & 0.712          & 2.54M    & 0.028s & \multicolumn{1}{c|}{0.747}          & 2.58M    & 0.036s & \textbf{0.762} \\
Cora     & SAGE-pool & 10     & 128                & 0.516          & 0.500          & 2.89M    & 0.060s & \multicolumn{1}{c|}{0.530}          & 3.05M    & 0.081s & \textbf{0.572} \\ \midrule
PubMed   & GCN       & 3      & 128                & 0.771          & 0.766          & 0.10M    & 0.027s & \multicolumn{1}{c|}{\textbf{0.780}} & 0.10M    & 0.035s & 0.771          \\
PubMed   & GCN       & 10     & 128                & 0.730          & 0.675          & 0.21M    & 0.068s  & \multicolumn{1}{c|}{0.719}          & 0.22M    & 0.090s & \textbf{0.740} \\
PubMed   & SAGE-mean & 3      & 128                & 0.795          & \textbf{0.797} & 0.20M    & 0.028s & \multicolumn{1}{c|}{0.791}          & 0.20M    & 0.036s & 0.785          \\
PubMed   & SAGE-mean & 10     & 128                & 0.750          & 0.754          & 0.43M    & 0.068s & \multicolumn{1}{c|}{0.756}          & 0.43M    & 0.083s & \textbf{0.765} \\
PubMed   & SAGE-pool & 3      & 128                & 0.753          & 0.759          & 0.50M    & 0.033s & \multicolumn{1}{c|}{\textbf{0.769}} & 0.53M    & 0.041s & 0.767          \\
PubMed   & SAGE-pool & 10     & 128                & 0.650          & 0.642          & 0.84M    & 0.079s & \multicolumn{1}{c|}{0.666}          & 1.00M    & 0.108s & \textbf{0.704} \\ \midrule
CiteSeer & GCN       & 3      & 128                & 0.633          & 0.617          & 0.51M    & 0.024s & \multicolumn{1}{c|}{0.607}          & 0.51M    & 0.030s & \textbf{0.647} \\
CiteSeer & GCN       & 10     & 128                & 0.249          & 0.253          & 0.62M    & 0.058s & \multicolumn{1}{c|}{0.486}          & 0.63M    & 0.080s & \textbf{0.553} \\
CiteSeer & SAGE-mean & 3      & 128                & 0.686          & 0.651          & 1.02M    & 0.028s & \multicolumn{1}{c|}{0.665}          & 1.02M    & 0.037s & \textbf{0.709} \\
CiteSeer & SAGE-mean & 10     & 128                & 0.429          & 0.454          & 1.25M    & 0.055s & \multicolumn{1}{c|}{0.418}          & 1.26M    & 0.078s & \textbf{0.579} \\
CiteSeer & SAGE-pool & 3      & 128                & 0.632          & 0.609          & 14,78M   & 0.044s & \multicolumn{1}{c|}{0.642}          & 14.82M   & 0.048s & \textbf{0.663} \\
CiteSeer & SAGE-pool & 10     & 128                & 0.436          & 0.397          & 15.13M   & 0.075s & \multicolumn{1}{c|}{0.491}          & 15.29M   & 0.080s & \textbf{0.512} \\ \bottomrule
\end{tabular}
}
\label{tab:cite}
\end{table*}

We conduct experiments with GCN as well as two variants of SAGE (SAGE-mean and SAGE-pool) on the Cora, PubMed, and CiteSeer datasets. The goal is to train these student models to classify unknown nodes on the visible graph. For GCN and two variants of SAGE, we adopt two versions of framework, one with 3 layers and the other with 10 layers. Both of them have 128 feature dimensions in each layer. Two student models with the same architecture are used for training. The results are shown in Table \ref{tab:cite}.

The three GNN models with 10 layers perform worse than those with 3 layers due to the existence of over-smoothing problem. For the GNN models with three layers, we find that Alignahead++ has limited enhancement in small datasets. In particular, the Alignahead++ method has few advantages in PubMed datasets, which is a very simple classification task and only needs to be divided into 3 classes, resulting in insufficient differentiation between various methods. When the number of layers increase to 10, the superiority of our Alignahead++ method becomes apparent. The Alignahead++ method significantly outperform the other methods on all datasets and all three GNN models. 

In addition, we plot the accuracy curves of GCN and SAGE-mean on Cora dataset when the number of layers increases gradually from 2 to 20, which is shown in Fig. \ref{fig:over-smoothing}. As the number of layers increases, the accuracy of all three methods decreases, but Alignahead++ decreases more slowly than other methods, which shows that Alignahead++ can effectively alleviate the over-smoothing problem in deep GNNs. Besides, Alignahead and Baseline become extremely unstable when deepening GNNs, that is, the accuracy curves fluctuate sharply, while the accuracy curve of Alignahead++ method is always flat.

\subsection{Sensitivity analysis}
\label{subsec:sensitivity}
In this section, we explore the impact of different values of $\beta$ and $\lambda$ on the results when fixing $\alpha=1$. The same experiment settings as the Section \ref{subsec:ppi} are used. The student model is a five-layer GAT, and the number of attention heads and feature dimensions of each layer are 2 and 128. We take 0, 0.2, 0.4, 0.6, 0.8 and 1 for $\beta$ and $\lambda$, respectively. The results are shown in Table \ref{tab:sen}.

When $\beta$ and $\lambda$ do not take extreme values (0 or 1), the performance of the model is relatively stable with around 0.9845 f1-score, demonstrating the robustness of our method. Besides, when both $\beta$ and $\lambda$ take 1, we find that the performance of the model deteriorates rapidly. Under the current hyper-parameter settings, the model can not obtain any label information referring to Table \ref{tab:extreme value}, which naturally leads to this result.
Furthermore, we find that the performance of the model doesn't seem to be affected when $\lambda=1$ and $\beta$ doesn't take the extreme values, that is, deep-supervised loss does not play a role here referring to Table \ref{tab:extreme value}. Next is our explanations. For shallow GNNs, each layer can capture the label information from the last layer by using Alignahead strategy when the over-smoothing problem is relatively light, making deep-supervised loss redundant. As the number of layers deepens, the over-smoothing problem becomes more serious, and the node features of the last layer become indistinguishable, making the intermediate layer cannot obtain the accurate label information. The huge error accumulated by back propagation degrades the model performance. At this time, adding the supervision of label in the intermediate layer can restrain the accumulation of error in the back propagation and guide the evolution of the intermediate layer to the right direction. 

\begin{table}[ht]
\centering
\renewcommand{\arraystretch}{1}
\caption{Sensitivity analysis of a shallow GAT on PPI dataset.}
\resizebox{0.8\linewidth}{!}{
\begin{tabular}{@{}c|cccccc@{}}
\toprule
\multicolumn{1}{c}{\diagbox{$\beta$}{$\lambda$}}   & 0      & 0.2    & 0.4    & 0.6    & 0.8    & 1      \\ \midrule
0   & 0.9810 & 0.9806 & 0.9806 & 0.9811 & 0.9805 & 0.9812 \\
0.2 & 0.9816 & 0.9847 & 0.9846 & 0.9848 & 0.9853 & 0.9849 \\
0.4 & 0.9823 & 0.9846 & 0.9843 & 0.9851 & 0.9856 & 0.9851 \\
0.6 & 0.9826 & 0.9842 & 0.9844 & 0.9853 & 0.9849 & 0.9845 \\
0.8 & 0.9815 & 0.9835 & 0.9841 & 0.9839 & 0.9838 & 0.9819 \\
1   & 0.9815 & 0.9825 & 0.9820 & 0.9816 & 0.9788 & 0.1907 \\ \bottomrule
\end{tabular}
}
\label{tab:sen}
\end{table}

\begin{figure}[ht]
  \centering
  \includegraphics[width=1\linewidth]{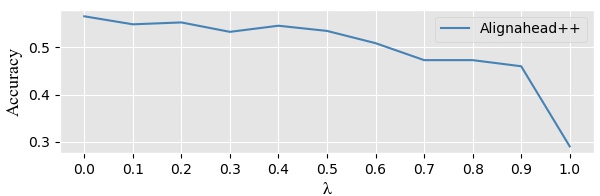}
  \caption{Fix $\beta=0.4$, the sensitivity analysis of hyper-parameter $\lambda$ on the CiteSeer dataset. The student model is 10-layer deep GCN with 128 feature dimensions.}
  \label{fig:lambda}
\end{figure}

Therefore, we perform sensitivity analysis on CiteSeer dataset and the student model is a 10-layer GCN with 128 feature dimensions. We fix $\beta=0.4$ and $\lambda$ was taken from 0 to 1 with an interval of 0.1. The result is shown in Fig. \ref{fig:lambda}. It is not difficult to find that with the increase of $\lambda$, the classification accuracy generally decreases, indicating that deep-supervised loss plays an important role in deep GNNs.

\subsection{Two different student models}
\label{subsec:difstu}
In addition to two identical student models, in this section we try to explore the performance of two different student models with the Alignahead++ method. Due to the limitations of Alignahead method, the two student models need to have the same number of layers. 
For the case of student model $S_{1}$ has $l^{S_{1}}$ layers and student model $S_{2}$ has $l^{S_{2}}$ layers, in which $l^{S_{1}}>l^{S_{2}}$, one feasible method is to arbitrarily choose $l^{S_{2}}$ layers from student model $S_{1}$ to align with student model $S_{2}$.
We select four groups of student models to conduct experiments on PPI dataset, in which all GAT models have 2 attention heads. Table \ref{tab:difstu} records the experimental results. We find that Alignahead++ method is superior to the competitors and the relatively poor student model tends to improve more in training. 

\begin{table}[ht]
\centering
\renewcommand{\arraystretch}{1}
\caption{Results with two different student models.}
\resizebox{1\linewidth}{!}{
\begin{tabular}{@{}cccc|c|c|c@{}}
\toprule
   & \multicolumn{3}{c|}{}             & Baseline    & Alignahead & Alignahead++ \\ \midrule
No & Model     & Layers & Feature dimensions & F1-score & F1-score   & F1-score      \\ \midrule
1  & GAT       & 4      & 128          & 0.9780   & 0.9814     & \textbf{0.9847}        \\
2  & SAGE-pool & 4      & 128          & 0.8885   & 0.9299     & \textbf{0.9487}        \\ \midrule
1  & GAT       & 5      & 128          & 0.9830   & 0.9837     & \textbf{0.9874}        \\
2  & SAGE-mean & 5      & 128          & 0.8941   & 0.9430     & \textbf{0.9514}        \\ \midrule
1  & GAT       & 5      & 256          & 0.9882   & 0.9887     & \textbf{0.9909}        \\
2  & GCN       & 5      & 256          & 0.5539   & 0.6128     & \textbf{0.6140}        \\ \midrule
1  & SAGE-mean & 10     & 512          & 0.9123   & 0.9896     & \textbf{0.9920}        \\
2  & GCN       & 10     & 512          & 0.4085   & 0.5056     & \textbf{0.6409}        \\ \bottomrule
\end{tabular}
}
\label{tab:difstu}
\end{table}

\subsection{The effect of auxiliary classifiers}
\label{subsec:aux}
The auxiliary classifiers, which are the key modules in Alignahead++, translate the unknown semantic space of the intermediate layers into the label semantic space. In CNNs, auxiliary classifiers are usually stacked by several convolutional layers, followed by one or two MLPs \cite{zhang2019byot,Yang2021hsakd,Xu2020sskd}. However, unlike images, graph data is irregular and carries structure information. Using MLPs directly as the auxiliary classifier is likely to lose structure information and affect model performance. We explore the differences between MLP and GNN as the auxiliary classifier and the effect of layer number on PPI dataset, which are shown in Table \ref{tab:classifier}. We use two GNN models, that is, four-layer SAGE-mean with 128 feature dimensions and six-layer GAT with 2 attention heads and 64 feature dimensions.

We find that GNN is indeed superior to MLP as an auxiliary classifier and increasing the number of auxiliary classifier layers does not always improve the student model performance. One possible reason is that more layer may cause the over-smoothing problem. Considering the training costs and model capacity, one layer of GNN auxiliary classifier is sufficient.

\begin{table}[ht]
\centering
\renewcommand{\arraystretch}{1}
\caption{The model performance with different auxiliary classifiers.}
\resizebox{0.5\linewidth}{!}{
\begin{tabular}{@{}ccc@{}}
\toprule
Auxiliary classifier & SAGE-mean & GAT \\ \midrule
1 MLP layer                  & 0.9323                              & 0.9570                                 \\
2 MLP layers                & 0.9360                               & 0.9552                               \\
1 GNN layer                  & 0.9455                                & 0.9590                                \\
2 GNN layers                & 0.9470                                & 0.9599                               \\
3 GNN layers                & 0.9404                               & 0.9598                               \\
4 GNN layers                & 0.9266                               & 0.9584                               \\ \bottomrule
\end{tabular}
}
\label{tab:classifier}
\end{table}

\subsection{The number of student models}
\label{subsec:stunumber}
In this section, we explore the effect of the number of student models on the experimental results. If the student number is greater than 2, each model will capture information from all remaining models. The $i$-th layer's structure and feature loss of the $k$-th model becomes
\begin{equation}
\begin{split}    
L_{l_{i}}^{S_{k}} = \frac{1}{M-1}\frac{1}{N}\sum_{h=1 \vee h\neq k}^{M}\sum_{j=1}^{N}D_{KL}\left(l_{i+1,j}^{S_{h}}||l_{i,j}^{S_{k}}\right) \\
L_{p_{i}}^{S_{k}} = \frac{1}{M-1}\frac{1}{N}\sum_{h=1 \vee h\neq k}^{M}\sum_{j=1}^{N}D_{KL}\left(p_{i+1,j}^{S_{h}}||p_{i,j}^{S_{k}}\right)
\end{split}
\end{equation}
where $M$ is the number of student models.
To make the student models have more room for improvement, we use the GAT with a smaller architecture on PPI dataset. The student model is a four-layer GAT, and the number of attention heads and feature dimensions of each layer are 2 and 42. We use the same parameter settings as in Section \ref{subsec:ppi}. 

\begin{table}[ht]
\centering
\renewcommand{\arraystretch}{1}
\caption{The effects of different student numbers on the PPI dataset. All student models are GAT with the same architecture. Metric is the max F1 score.}
\resizebox{0.8\linewidth}{!}{%
\begin{tabular}{@{}c|ccccc@{}}
\toprule
Number of models & 2      & 3      & 4      & 5      & 6      \\ \midrule
OC               & 0.8482 & 0.8505 & 0.8498 & 0.8504 & 0.8507 \\
Alignahead       & 0.8502 & 0.8511 & 0.8523 & 0.8526 & 0.8525 \\
Alignahead++    & 0.8540 & 0.8542 & 0.8545 & 0.8541 & 0.8539 \\ \bottomrule
\end{tabular}
}
\label{tab:model-number}
\end{table}

As shown in Table \ref{tab:model-number}, increasing the student numbers does improve the model performance especially when it is less than 4 in Alignahead. For Alignahead++, however, two student models are sufficient. When the number increases further, the Max F1 score hardly changes, which indicates that the model performance has reached its limit. 
Since the model training time linearly increases as the student numbers increase, two or three student models are generally appropriate.

\section{Conclusion}
In this paper, we propose an online knowledge distillation framework specifically for graph neural networks, where two student models extract information from each other in an alternating training procedure. Specifically, we design a strategy called Alignahead to align one student layer with the next layer of another student model, resulting in the information spreading over all layers after several iterations. We further develop a new framework with our existing approach, namely Alignahead++, which introduces feature information except structure information by adding auxiliary classifier in each intermediate layer, to alleviate the over-smoothing problem in deep GNNs with deep supervision. We conduct experiments with GAT, GCN and two variants of SAGE on four datasets: PPI, Cora, PubMed and CiteSeer. The results show that the new proposed Alignahead++ is superior to Alignahead and other compared methods, especially in deep GNNs. Beyond that, we find that increasing the student numbers generally improves the performance further and our framework exhibits high robustness with different hyper-parameter settings. 

In the future, we will consider student models with different layer numbers, other GNN models and datasets with different properties to demonstrate the applicability and effectiveness of our proposed framework. In addition, since not all information from different layers has the same effect on the final task, we are interested to explore appropriate information weighting method for better training. 

\section{Acknowledgment}
This work is supported by the Starry Night Science Fund of Zhejiang University Shanghai Institute for Advanced Study (Grant No: SN-ZJU-SIAS-001) and National Natural Science Foundation of China (Grant No: U1866602). The authors would like to thank anonymous reviewers for their helpful comments.
\bibliographystyle{IEEEtran}
\bibliography{egbib}


\end{document}